\documentclass[final]{cvpr}

\usepackage{times}
\usepackage{enumitem}
\usepackage{epsfig}
\usepackage{graphicx}
\usepackage{amsmath}
\usepackage{amssymb}
\usepackage{indentfirst}
\usepackage{algorithm}
\usepackage{algorithmic}
\usepackage{multirow}
\usepackage{indentfirst}
\usepackage{booktabs}

\usepackage{latexsym}
\usepackage{array}
\newcommand{\PreserveBackslash}[1]{\let\temp=\\#1\let\\=\temp}
\newcolumntype{C}[1]{>{\PreserveBackslash\centering}p{#1}}
\newcolumntype{R}[1]{>{\PreserveBackslash\raggedleft}p{#1}}
\newcolumntype{L}[1]{>{\PreserveBackslash\raggedright}p{#1}}

\usepackage[pagebackref=true,breaklinks=true,colorlinks,bookmarks=false]{hyperref}

\begin{document}

\title{Point-Level Temporal Action Localization: \\Bridging Fully-supervised Proposals to Weakly-supervised Losses}

\author{Chen Ju\\
Shanghai Jiao Tong University\\
{\tt\small ju\_chen@sjtu.edu.cn}
\and
Peisen Zhao\\
Shanghai Jiao Tong University\\
{\tt\small pszhao@sjtu.edu.cn}
\and
Ya Zhang\\
Shanghai Jiao Tong University\\
{\tt\small ya\_zhang@sjtu.edu.cn}
\and
Yanfeng Wang\\
Shanghai Jiao Tong University\\
{\tt\small wangyanfeng@sjtu.edu.cn}
\and
Qi Tian\\
Huawei Cloud \& AI\\
{\tt\small tian.qi1@huawei.com}
}

\maketitle

\begin{abstract}
Point-Level temporal action localization (PTAL) aims to localize actions in untrimmed videos with only one timestamp annotation for each action instance. Existing methods adopt the frame-level prediction paradigm to learn from the sparse single-frame labels. However, such a framework inevitably suffers from a large solution space. This paper attempts to explore the proposal-based prediction paradigm for point-level annotations, which has the advantage of more constrained solution space and consistent predictions among neighboring frames. The point-level annotations are first used as the keypoint supervision to train a keypoint detector. At the location prediction stage, a simple but effective mapper module, which enables back-propagation of training errors, is then introduced to bridge the fully-supervised framework with weak supervision. To our best of knowledge, this is the first work to leverage the fully-supervised paradigm for the point-level setting. Experiments on THUMOS14, BEOID, and GTEA verify the effectiveness of our proposed method both quantitatively and qualitatively, and demonstrate that our method outperforms state-of-the-art methods.
\end{abstract}

\section{Introduction}    \label{introduction}
Temporal action localization (TAL), which localizes actions from untrimmed videos, plays an important role in video understanding. Recently, the fully-supervised setting has achieved impressive results~\cite{chao2018rethinking,lin2018bsn,lin2019bmn,shou2016temporal,gao2017turn,lin2019fast,zhao2017temporal}, but the precise action boundary annotations are time-consuming and hence expensive. The video-level weak-supervised setting~\cite{liu2019completeness,nguyen2019weakly,lee2019background,paul2018w,shou2018autoloc,liu2019weakly} only requires cheaper category labels for localization, but the lack of explicit location guidance limits its performance highly inferior to the fully-supervised counterpart.

\begin{figure}[t]
\begin{center}
\vspace{5pt}
\includegraphics [width=0.476\textwidth] {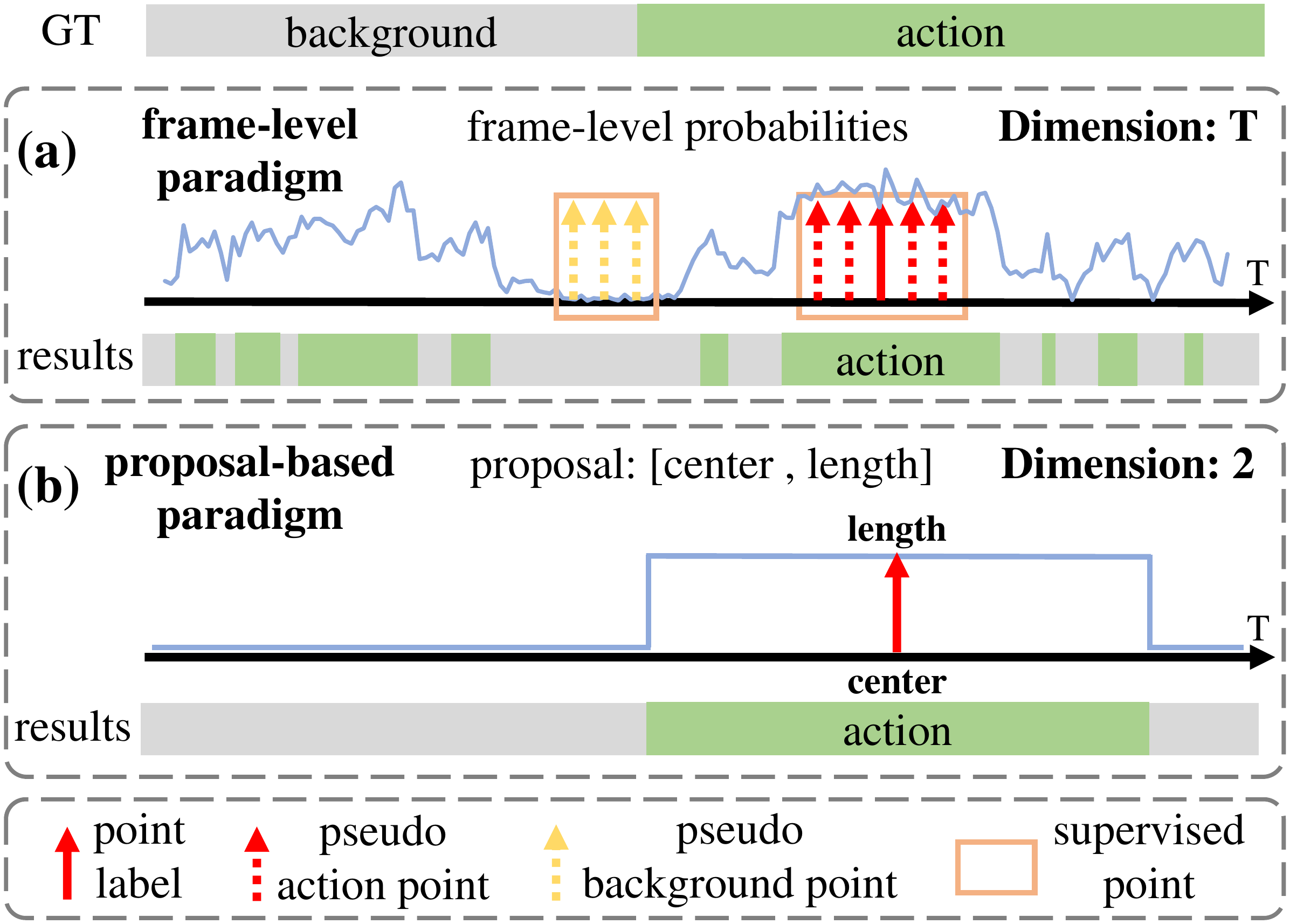}
\end{center}
\caption{Comparison of two prediction paradigms. \textit{(a): the frame-level paradigm} uses partial pseudo labels to supervise the action probability for each frame. \textit{(b): the proposal-based paradigm} directly predicts the action center and action length, and has a smaller solution space.}
\label{fig:intro}
\end{figure}

To bridge the performance gap between fully-supervised setting and video-level weakly-supervised setting while maintaining low annotation overhead, a point-level supervision setting (abbreviated as PTAL) is introduced~\cite{ma2020sf}, which provides a single frame (point) annotation for each action instance during training. To reduce the label sparsity associated with the single-frame annotation, pseudo label mining is employed to increase the number of labeled action frames and background frames through a self-training like expansion strategy. However, the labels obtained from such expansion strategy are usually incomplete and imprecise due to the trade-offs between the quality and quantity of such pseudo labels. To learn from these pseudo labels, SF-Net~\cite{ma2020sf} adopts the frame-level prediction paradigm as illustrated in Figure~\ref{fig:intro} (a), which enables each frame to independently make a prediction and independently be evaluated. Such a framework inevitably suffers from a large solution space, similar to other weakly-supervised frameworks~\cite{paul2018w,lee2019background,liu2019completeness,narayan20193c}. As a result, predicting frame-level probabilities with such pseudo-labeled frames has led to high false-positive rates and discontinuous actions.

In this paper, we attempt to explore the proposal-based prediction paradigm~\cite{shou2016temporal,shou2017cdc,gao2017turn,yang2020revisiting} widely adopted in the fully-supervised setting for PTAL. As shown in Figure~\ref{fig:intro} (b), instead of predicting frame-level probabilities, the proposal-based paradigm generates proposals based on anchor points to represent action instances. The proposal location is restricted near the anchor point, and the action probabilities of frames within the same proposal are naturally constrained to be consistent, thus greatly reducing the solution space. Following the proposal-based paradigm, we regard the point-level annotations as the keypoint supervision, then train a keypoint detector to identify action anchor points as the rough location of each action instance. For each anchor point, similar to the fully-supervised setting, an action proposal in terms of the action center and its corresponding length is directly predicted. However, it is non-trivial to supervise the prediction of action proposals with category weak labels due to the lack of boundary locations.

A straight-forward choice is to turn the fully-supervised proposal, \emph{i.e.}, action center and its length, into the proposal-level classification probability, so that the supervision in terms of category labels can be leveraged. Inspired by the attention mechanism~\cite{vaswani2017attention}, we propose to transform the proposal location into a binary temporal mask which is further used as temporal attention to obtain the proposal-level classification probability. While the mapping between the location of the action proposal and the mask seems mathematically simple, it is unfortunate that such a direct transformation function is non-differentiable. To back-propagate training errors for optimization, we further design a simple but novel mapper module, which learns the transformation from simulated data. Through the above design, we bridge fully-supervised proposals to weakly-supervised labels. To our best of knowledge, this is the first work to leverage the fully-supervised paradigm for the point-level setting.

On three benchmark datasets, BEOID~\cite{damen2014you}, GTEA~\cite{lei2018temporal} and THUMOS14~\cite{jiang2014thumos}, our method outperforms previous state-of-the-art methods, both quantitatively and qualitatively. We further perform extensive ablation analyses and comparisons to reveal the effectiveness of each component. In summary, our contributions are as follows.
(1) We introduce the proposal-based prediction paradigm to the point-level supervision setting for temporal action localization, which greatly reduces the issues of discontinuous action predictions and false positives.
(2) We propose a simple and effective mapper to bridge fully-supervised proposals to weak supervision, hence enabling the paradigm to be compatible with weak and full supervision.

\begin{figure*}[t]
\begin{center}
\vspace{5pt}
\includegraphics [width=0.95\textwidth] {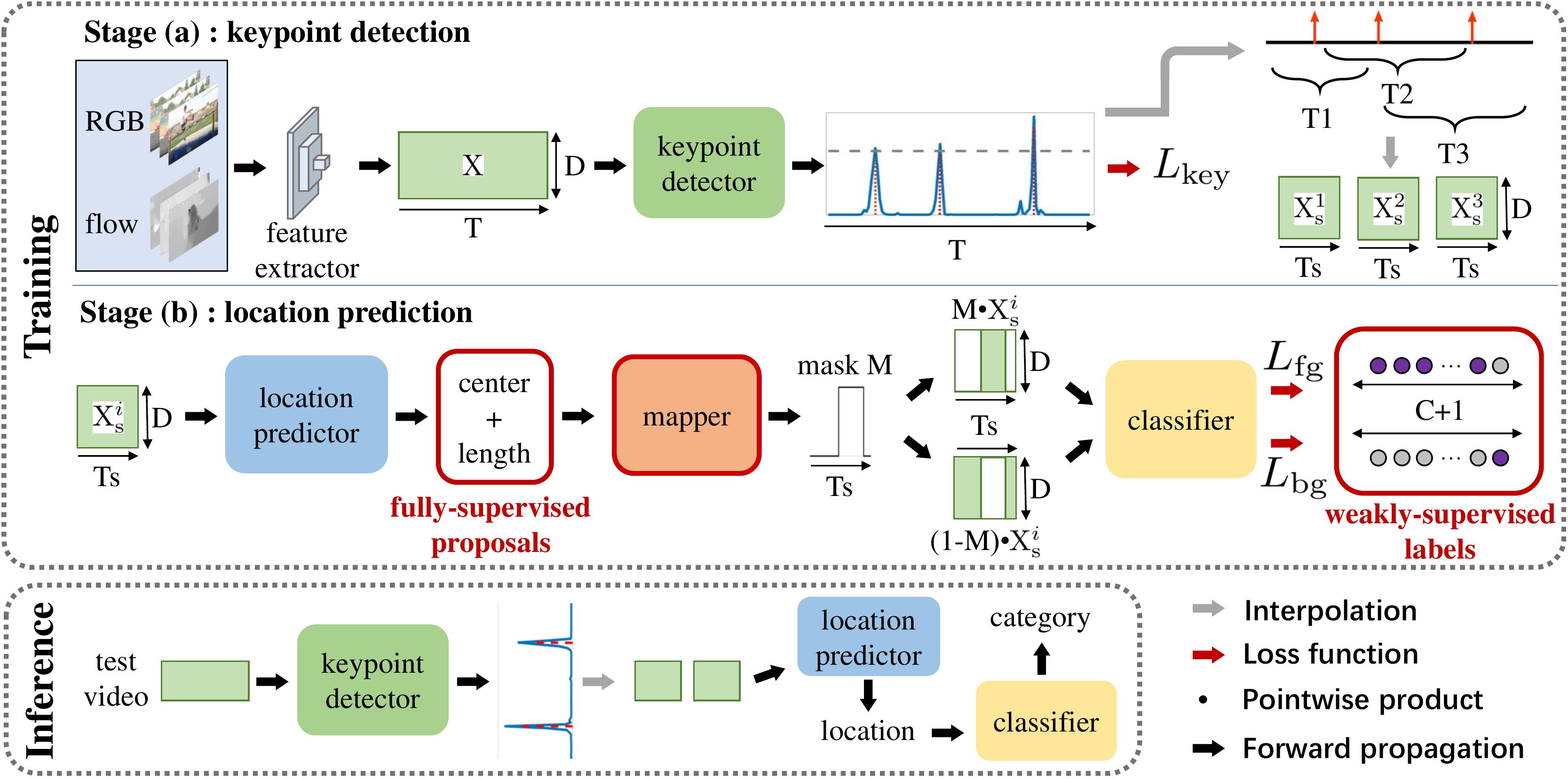}
\end{center}
\vspace{-5pt}
\caption{Framework pipeline. The model is trained in two sequential stages. In stage (a), the keypoint detector is trained via point-level annotations to evaluate the key probability of each timestamp. The peaks in the key heatmap are regarded as anchor points to segment the video into several short videos. In stage (b), the location predictor outputs a proposal for each short video. Then the pre-trained mapper is frozen to transform the proposal location into a binary temporal mask. The resulting masked features are fed into the classifier to construct loss functions with weakly-supervised category labels.}
\label{fig:framework}
\end{figure*}


\section{Related Work}
\textbf{Fully-supervised temporal action localization}, which requires precise action boundary annotations, has made great progress recently. The popular solution is to generate proposals representing action instances first, then classify them. There are two main paradigms for generating proposals, namely top-down framework~\cite{shou2016temporal,shou2017cdc,gao2017turn,chao2018rethinking} and bottom-up framework~\cite{zhao2017temporal,lin2018bsn,lin2019bmn,lin2019fast}.
The former generates sufficient anchor points on the video through sliding windows, and produces a proposal for each anchor point based on the preset length. A regressor is utilized to adjust the proposal boundaries.
The latter trains a detector to search extreme points (such as boundary points, center points) as action anchor points, then combines extreme points or performs length predictions to generates proposals.
Both kinds of proposal-based paradigms employ anchor points to identify the rough locations of action instances, and constrain the consistency within proposals, thus greatly reducing the solution space.
However, since all these methods require huge annotation costs, they are not compatible with weak supervision settings and cannot be widely used in reality.

\textbf{Weakly-supervised temporal action localization} is proposed to reduce high annotation costs.
The most widely used is video-level category labels, and the corresponding methods are divided into two branches. The MIL-based framework~\cite{paul2018w,lee2019background,liu2019completeness,narayan20193c} first trains a video-level classifier, then obtains frame-level action probabilities by checking the produced Class Activation Sequence (CAS). The Attention-based framework~\cite{nguyen2019weakly,nguyen2018weakly,shou2018autoloc,liu2019weakly} directly predicts frame-level action probabilities from raw data, which is regarded as attention to calculate video-level classification probabilities for model optimization. Besides, the number of action instances~\cite{narayan20193c,xu2018segregated} has also been explored to provide more action information.
Nevertheless, all above methods rely on insufficient category or count labels to predict frame-level action probabilities and require the empirically preset threshold to post-process the probabilities for localization results. Therefore, they are all troubled by serious background false positives and incomplete action predictions~\cite{nguyen2019weakly,lee2019background,liu2019completeness}, which lead to a huge performance gap between them and fully-supervised methods.

\textbf{Point-level supervision} is widely used to balance labeling costs and model performance.
In image semantic segmentation task, WTP~\cite{bearman2016s} introduced the point-level setting by annotating a single pixel for each instance. PDML~\cite{qian2019weakly} followed this setting and took advantage of these point labels for metric learning. In object counting task, CLPS~\cite{laradji2018blobs} designed novel split-level loss and false-positive loss based on point-level annotations.
In video tasks, marking one spatial location in the frame for each instance was first proposed by SPOT~\cite{mettes2016spot} to improve spatial-temporal localization. ARST~\cite{moltisanti2019action} annotated a single timestamp of each action instance for action recognition. SF-Net~\cite{ma2020sf} extended it to temporal action localization and obtained certain improvements.
However, they use insufficient labels to predict the action probability for each frame, causing serious discontinuous action predictions and background false positives.
On the contrary, we introduce the proposal-based paradigm to constrain and reduce the solution space, and design a novel mapper to bridge the paradigm to category weak labels, effectively tackling their issues.

\section{Approach}
We set point-level supervision following~\cite{ma2020sf}: for each action instance in $N$ untrimmed training videos, we provide it with one timestamp $t$ and action category $\mathbf{y}$, where $\mathbf{y} \in\mathbb{R}^{C}$ and $C$ is the total number of action categories. For each testing video, we are expected to predict a set of action instances in the form of the start time, the end time and the action category. Notably, each video can contain multiple categories and multiple action instances.

Due to the great variation in video lengths, we first sample $T$ consecutive snippets from each video, then generate the RGB and flow features using the pre-trained extractor. By concatenating two-stream features, we obtain the video feature map $\mathbf{X}\in\mathbb{R}^{T\times{D}}$, where $D$ is the feature dimension.

\subsection{Overview}
The absence of complete labels in PTAL makes the popular frame-level probability prediction paradigm confused in a large solution space, causing serious discontinuous action predictions and false positives.
To solve these issues, we introduce the proposal-based prediction paradigm in full supervision, to add more prediction constraints and reduce the solution space.
As illustrated in Figure~\ref{fig:framework}, to identify the rough locations of action instances, we first generate some action anchor points via a keypoint detector supervised by point-level annotations \textbf{(stage (a))}. Each anchor point indicates an action instance, thereby limiting action regions and eliminating false positives. Then for each anchor point, we predict the action center and action length to form a proposal \textbf{(stage (b))}, ensuring the continuity of predictions.
To optimize this fully-supervised paradigm with weakly-supervised labels, we further design a novel mapper module to transform the proposal location into a binary temporal mask, and sequentially classify the masked videos to construct supervision with category labels.

\subsection{Keypoint Detection}
\textbf{Keypoint detector.}
The goal of keypoint detector is to identify the rough action locations through some anchor points, thereby eliminating background false positives. However, there are no precise location labels, but only point-level annotations available in PTAL. Fortunately, we observe that annotators tend to point out the discriminative moments for action instances. Hence, these annotated points are keypoints of actions, providing explicit guidance for distinguishing discriminative action points from the background, even if they are similar in appearance.

With such point-level annotations, we learn a keypoint detector (implemented by a fully convolutional network) to evaluate the key probability of each timestamp in the video. The detector is fed with video features $\mathbf{X}$, and output keypoint estimate heatmap $\mathbf{\hat{K}}\in\mathbb{R}^{T\times{C}}$.

For the training labels, if a video frame is selected as the annotation point, it is regarded as a positive sample; otherwise, it is treated as a negative sample.
And we follow~\cite{lin2018bsn,zhao2020bottom,lin2019bmn} to define the keypoint loss with the weighted cross-entropy:
\begin{equation}
    {L_{\mathrm{key}} = \frac{1}{T^{+}} \sum _{t\in \Omega^{+}} \mathcal{H}(\mathbf{k}_t, \mathbf{\hat{k}}_t) +  \frac{1}{T^{-}} \sum _{t\in \Omega^{-}} \mathcal{H}(\mathbf{k}_t, \mathbf{\hat{k}}_t),}
\end{equation}
where ${\mathbf{k}_t} \in\mathbb{R}^{{C}}$ and ${\mathbf{\hat{k}}_t} \in\mathbb{R}^{{C}}$ are the ground truth and estimated key probability of the $t$-th frame, $\mathcal{H}$ denotes the regular cross-entropy loss, $\Omega^{+}$ and $\Omega^{-}$ mean the positive and negative sample sets, $T^{+}$ and $T^{-}$ mean the number of positive and negative samples, respectively.

\textbf{Keypoint generation.}
From the keypoint heatmap, we get anchor points by mining local maximum points.
For any timestamp, we regard it as a keypoint if its key probability peaks in the heatmap and exceeds the set threshold $\theta$. The filtered timestamps are then sorted and grouped into a candidate keypoint set $\mathcal{P}=\{p_j\}_{j=1}^{N_p}$, where $N_p$ is the number of keypoints.
Each keypoint correspondings to a unique action instance, and there is the background between any two keypoints to divide the corresponding instances.

Due to the absence of precise location labels, it is difficult to process videos with a different number of action instances. Since keypoints naturally divide different instances, we propose to separate the entire video into several short videos based on the keypoint set $\mathcal{P}$, to ensure that each short video only contains a complete instance.
Formally, for the $j$-th keypoint $p_j$ in the set, we set the temporal interval of its corresponding short video as $[{p_{j-1}+1},{p_{j+1}-1}]$.
To fix the temporal length, we rescale each short video to $T_{\mathrm{s}}$ frames by linear interpolation.
Next, we predict the action location for each short video.

\subsection{Location Prediction}  \label{sec:Length Prediction}
In the location prediction stage, we expect to predict an action proposal for each short video, and apply category labels for optimization. Concretely, we generate the proposal with a location predictor, then utilize a novel mapper to transform the proposal location into a binary temporal mask. Regarding the mask as weights, we calculate the foreground action features and background features, and finally classify them with a classifier.

\textbf{Location predictor.}
The goal of location predictor is to generate proposals indicating action locations based on the keypoints, to ensure the continuity of predictions.
As we are not confident whether the keypoint is at the center of the action, we form a proposal by the action length and the offset between the keypoint and the center point.
Formally, we input the raw features $\mathbf{X}_{\mathrm{s}}\in\mathbb{R}^{{T_{\mathrm{s}}}\times{D}}$ of the short video into the location predictor, and obtain the proposal $\mathbf{v} = [{\Delta}p+p, l]$, where $p$, ${\Delta}p$, $l$ represent the keypoint location, the center offset and the action length.
The temporal boundaries of the proposal are given by:
\begin{equation}  \label{eq:actionboundary}
    {r_a = {\Delta}p+p-\frac{l}{2}  \quad  r_b = {\Delta}p+p+\frac{l}{2},}
\end{equation}
where $r_a$ and $r_b$ denote the left and right boundaries.


\begin{figure}[t]
\begin{center}
\includegraphics [width=0.48\textwidth] {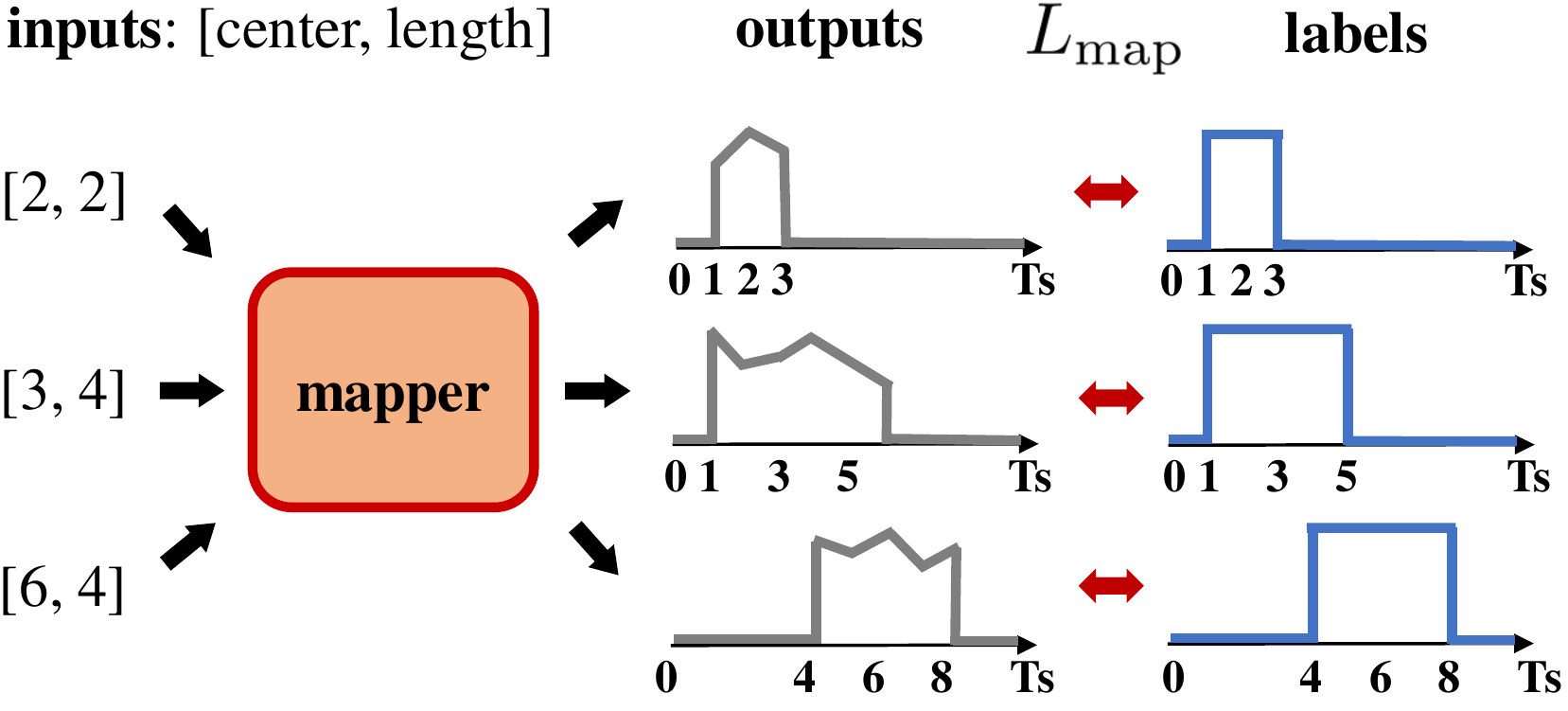}
\end{center}
\caption{Paired training data of the mapper. The input is a two-dimensional proposal, indicating the action center and action length. The label is a $T_{\mathrm{s}}$-dimensional binary mask, indicating the temporal interval of the proposal.
}
\label{fig:mapper}
\end{figure}


\textbf{Mapper.}   \label{mapper}
To optimize the location predictor only with classification supervision, we are urgent to calculate the proposal-level category probability according to the proposal $\mathbf{v}$.
Inspired by the attention mechanism~\cite{vaswani2017attention}, we propose to transform the proposal location into a binary temporal mask $ \mathbf{m} \in \{0,1\}^{T_{\mathrm{s}}}$, then produce the proposal-level features through the mask for subsequent classification.
Although mathematically simple, such a direct transformation is non-differentiable, which makes it infeasible to back-propagate training errors for model optimization.

To tackle this, we leverage a simple but effective mapper module (constructed by the Multi-Layer Perceptron) to fit this transformation in advance, then freeze the pre-trained weights of the mapper during the training of location prediction stage, thus ensuring the accurate transformation and maintaining the error back-propagation.
To train the mapper, all we need is to generate enough paired data defined in Figure~\ref{fig:mapper}.
We first randomly generate lots of simulated proposals, representing action centers and action lengths respectively, as the input data.
Then for each proposal, we define a $T_{\mathrm{s}}$-dimensional binary temporal mask as its ground truth label.
In specific, we assign the positive foreground for all frames whose temporal locations are inside the proposal interval, and the {negative background} for all frames whose temporal locations are outside the interval.
The binary mask ${m}_t$ of $t$-th frame is formalized as follows:
\begin{equation}
{m}_t = \left\{ \begin{aligned}
1, & & \mathrm{if} \ t \in [r_a,r_b] \\
0, & & \mathrm{if} \ t \notin [r_a,r_b] \\
\end{aligned} \right.
\end{equation}

We also employ the weighted cross-entropy loss to optimize the mapper module.
\begin{equation}
    {L_{\mathrm{map}} = \frac{1}{T_{\mathrm{s}}^{+}} \sum _{t\in \Lambda^{+}} \mathcal{H}(m_t, \hat{m}_t) +  \frac{1}{T_{\mathrm{s}}^{-}} \sum _{t\in \Lambda^{-}} \mathcal{H}(m_t, \hat{m}_t)}
\end{equation}
where $\hat{m}_t$ is the mapper output of the $t$-th frame, $\mathcal{H}$ is the regular cross-entropy loss, $\Lambda^{+}$ and $\Lambda^{-}$ denote the positive and negative sample sets, $T^{+}_s$ and $T^{-}_s$ are the number of positive and negative samples.

\textbf{Classifier.}
In this module, we perform foreground classification and background modeling to better distinguish foreground actions and background.
As ideal attention, the binary mask $\mathbf{m} \in \{0,1\}^{T_{\mathrm{s}}}$ excellently identifies the foreground location. Hence, we regard it as weights to filter out all the foreground action features, then perform average pooling over these filtered features to calculate the proposal-level foreground feature $\mathbf{x}_{\mathrm{fg}}\in\mathbb{R}^{D}$:
\begin{equation}
    {\mathbf{x}_{\mathrm{fg}} = \frac{1}{T_{\mathrm{s}}}\sum_{t=1}^{T_{\mathrm{s}}}{m_t}\mathbf{x}_t.}
\end{equation}

Similarly, the complement weights $\mathbf{1-m}$ can be utilized to indicate the background location, and the proposal-level background feature $\mathbf{x}_{\mathrm{bg}}\in\mathbb{R}^{D}$ is obtained by:
\begin{equation}
    {\mathbf{x}_{\mathrm{bg}} = \frac{1}{T_{\mathrm{s}}}\sum_{t=1}^{T_{\mathrm{s}}}{(1-m_t)}\mathbf{x}_t,}
\end{equation}
where $\mathbf{x}_t \in\mathbb{R}^{D}$ is the short video feature of $t$-th frame.
After that, we input these two features into the same classifier (built by the Multi-Layer Perceptron) to predict the foreground classification probability $\hat{y}_{\mathrm{fg}}$ and the background aware probability $\hat{y}_{\mathrm{bg}}$. Note that we adopt $C$ action categories and one background category for classification, hence $\hat{y}_{\mathrm{fg}} \in\mathbb{R}^{C+1}$ and $\hat{y}_{\mathrm{bg}} \in\mathbb{R}^{C+1}$.

To optimize the classifier, we apply the regular cross-entropy loss between the predicted category probabilities and the corresponding ground truth category labels:
\begin{equation}  \label{beta}
    {L_{\mathrm{cls}} =L_{\mathrm{bg}} + \beta L_{\mathrm{fg}}= \mathcal{H}(\mathbf{y}_{\mathrm{bg}}, \mathbf{\hat{y}}_{\mathrm{bg}}) + \beta \sum_{c=1}^C \mathcal{H}(y_{\mathrm{fg}}^c, \hat{y}_{\mathrm{fg}}^c) ,}
\end{equation}
where $\mathcal{H}$ means the regular cross-entropy loss, $\beta$ is a trade-off hyperparameter, $\mathbf{y_{\mathrm{fg}}}=[y^1, ..., y^C, 0]^\mathrm{T} \in\mathbb{R}^{C+1}$ and $\mathbf{y_{\mathrm{bg}}}=[0, ..., 0, 1]^\mathrm{T} \in\mathbb{R}^{C+1}$ are the foreground and background category labels, respectively.

\subsection{Inference}
At testing time, different from previous methods~\cite{ma2020sf,lee2019background,shi2020weakly,nguyen2019weakly}, our method is elegant without the need for post-processing, \emph{e.g.}, non-maxima suppression (NMS).
For a given video, we first utilize the keypoint detector to predict its key heatmap, then extract the peaks in the heatmap as action anchor points. Based on these points, we divide the entire video into several short videos, each containing only one keypoint.
Afterward, for each short video, we feed it into the location predictor to calculate the start time and the end time of the action proposal, and use the classifier to predict the proposal category. Each proposal is scored with the corresponding keypoint probability.

\section{Experiments}
\subsection{Datasets and Evaluation}
We conduct experiments on the following three datasets. For the sake of fairness, we adopt the single-frame annotations labeled in~\cite{ma2020sf} to provide point-level location supervision for each action instance during training.

\textbf{THUMOS14}~\cite{jiang2014thumos} contains 413 untrimmed sports videos, which belong to 20 action categories. Following the convention, we train on 200 validation videos and evaluate on 213 testing videos. There are 3007 point-level annotations available for training and each video contains an average of 15 action instances. Besides, the actions and videos vary widely in length, making this dataset particularly challenging.
\textbf{BEOID}~\cite{damen2014you} covers 58 videos in 34 action categories. According to~\cite{ma2020sf,moltisanti2019action}, we set the proportion of training and testing videos to 80-20\%, and obtain a total of 594 point-level annotations.
\textbf{GTEA}~\cite{lei2018temporal} records 7 fine-grained actions in the kitchen. There are 28 videos in total, divided into 21 videos for training and 7 videos for testing. Each training video contains 17.5 point-level labels on average.

\textbf{Evaluation Metrics.}
We follow the standard protocols to evaluate with mean Average Precision (mAP) under different intersection over union (IoU) thresholds. And a proposal is regarded as positive only if both IoU exceeds the set threshold and the category prediction is correct.


\begin{table}[t]
\small
\begin{center}
\caption{Evaluation of core modules on THUMOS14. `Key' indicates whether to detect keypoints. Two location prediction paradigms: frame-level probability and proposal-based prediction are abbreviated as `frame' and `proposal'. AVG is the average mAP at IoU thresholds 0.1:0.1:0.7.
}
\setlength{\tabcolsep}{5.5pt}{\begin{tabular}{c|c|c|ccc|c}
\toprule
\multirow{2}{*}{Setting} & \multirow{2}{*}{Key} & \multirow{2}{*}{Paradigm} & \multicolumn{3}{c|}{mAP@IoU} & \multirow{2}{*}{AVG} \\ \cline{4-6}
 &  &  & 0.3 & 0.5 & 0.7 &  \\ \hline \hline
\multirow{3}{*}{Point} & no & frame & 51.7 & 29.3 & 9.2 & 39.6  \\
 & yes & frame & 55.2 & 30.7  & 9.8 & 41.7  \\
 & yes & proposal & 58.1 & 34.5 & 11.9 & 44.3 \\ \hline
Full & yes & proposal & \textbf{60.1} & \textbf{39.2} & \textbf{18.4} & \textbf{48.3} \\ \bottomrule
\end{tabular}}
\label{tab:mil vs 2dvector}
\end{center}
\end{table}


\subsection{Implementation Details}
\textbf{Feature Extraction.}
Following previous literature~\cite{paul2018w,liu2019completeness,nguyen2019weakly,liu2019weakly}, we first split each untrimmed video into 16-frame non-overlapping snippets, then extract the optical flow from RGB data via the TV-L1 algorithm~\cite{wedel2009improved}. For fair comparison, we adopt the classic two-stream I3D~\cite{carreira2017quo} network as the feature extractor and fix the parameters pre-trained on Kinetics dataset~\cite{carreira2017quo}. After obtaining RGB and flow features, we integrate them in an early-fusion fashion, and get a 2048-dimensional vector for each snippet. The number of snippets $T$ is fixed to 2500, 360 and 128 for THUMOS14, BEOID and GTEA, respectively.

\vspace{5pt}
\textbf{Parameter settings.}
For all datasets, our method is trained by the Adam optimizer~\cite{kingma2014adam} with a learning rate of $10^{-4}$. To train the mapper, we simulate one million paired data and employ the Adam optimizer with a learning rate of $10^{-5}$. For the trade-off hyperparameter $\beta$ in Eq.~\ref{beta}, we set it to 1.25 on THUMOS14 and BEOID, and 2 on GTEA. And the threshold $\theta$ in keypoint generation is set to 0, 0.01 and 0.15 for GTEA, BEOID and THUMOS14, respectively. To eliminate the high-frequency noise in the keypoint heatmap, we employ the Savitzky-Golay filter~\cite{press1990savitzky} for smoothing. The specific network architectures and more details are reported in the supplementary material.


\begin{table}[t]
\small
\begin{center}
\caption{Ablation studies of the location prediction stage on THUMOS14. ${\Delta}p$ is the center offset, $L_{\mathrm{fg}}$ and $L_{\mathrm{bg}}$ are the foreground and background losses in Eq.~\ref{beta}.
AVG means the average mAP at IoU thresholds 0.1:0.1:0.7.
}
\begin{tabular}{ccc|ccc|c}
\toprule
\multirow{2}{*}{{$L_{\mathrm{fg}}$}} & \multirow{2}{*}{{$L_{\mathrm{bg}}$}} & \multirow{2}{*}{{${\Delta}p$}} & \multicolumn{3}{c|}{{mAP@IoU}} & \multirow{2}{*}{{AVG}} \\ \cline{4-6}
 &  &  & {0.3} & {0.5} & {0.7} &  \\ \hline \hline
\checkmark &  &  & 51.9 & 27.2 & 8.0 & 38.7 \\
\checkmark & \checkmark &  & 57.1 & 33.8 & 11.5 & 43.8  \\
\checkmark &  & \checkmark & 53.0 & 28.1 & 8.7 & 39.6  \\
\checkmark & \checkmark & \checkmark &\textbf{58.1} & \textbf{34.5} & \textbf{11.9}  & \textbf{44.3} \\ \bottomrule
\end{tabular}
\label{tab:ablation studies}
\end{center}
\vspace{-10pt}
\end{table}


\vspace{10pt}
\subsection{Evaluation of Core Modules}
In this section, we evaluate the effectiveness of keypoint detection and proposal-based location prediction.
Firstly, the baseline is set to predict the action probability for each frame through point-level annotations without keypoint detection. We adopt the frame classification loss and video classification loss in~\cite{ma2020sf} for optimization, and use the classic threshold post-processing~\cite{ma2020sf,shi2020weakly,liu2019completeness} to produce the final predictions. Secondly, we add the keypoint detection to the baseline, and only retain the predictions containing keypoints. For the predictions with multiple keypoints, we divide them by regarding keypoints as boundaries.
Finally, we continue to retain the keypoint detection and replace the frame-level probability prediction paradigm with the proposal-based location prediction paradigm, to form our complete method.
Note that the proposal-based prediction paradigm cannot stand alone without keypoints, as keypoints act as anchor points for actions.

The mAP results are summarized in Table~\ref{tab:mil vs 2dvector}.
We can obverse that the baseline performs worst among these three methods. And with the help of keypoint detection, there yields a stable increase in performance, with a gain of 2.1\% on the average mAP. This observation indicates that detecting keypoints as action anchor points indeed eliminates many false positives.
With anchor points, the detector limits action localization regions, thus greatly reducing the solution space and suppressing background activation.
Moreover, even if only supervised by classification, the proposal-based prediction paradigm outperforms the frame-level probability paradigm by a large margin. We attribute the improvement to more complete action predictions. By directly generating the action length, the proposal-based paradigm naturally ensures the continuity of predictions, thereby further constraining the solution space and stably outputting high-quality predictions.
Furthermore, to explore the performance upper bound of our method, we replace point-level labels with precise boundary labels to evolve into a fully-supervised setting. There emerge significant boosts in performance, especially at high IoU thresholds, revealing the wide compatibility of our method.

\begin{table}[t]
\small
\begin{center}
\caption{Comparison of different label distributions on THUMOS14. AVG denotes the average mAP at IoU thresholds 0.1:0.1:0.7. The labels are generated by simulation.}
\setlength{\tabcolsep}{5.5pt}{\begin{tabular}{c|c|ccc|c}
\toprule
\multirow{2}{*}{{Method}} & \multirow{2}{*}{{Distribution}} & \multicolumn{3}{c|}{{mAP@IoU}} & \multirow{2}{*}{{AVG}} \\ \cline{3-5}
 &  & {0.3} & {0.5} & {0.7} &  \\ \hline \hline
\multirow{3}{*}{SF-Net~\cite{ma2020sf}}
 & Manual & 53.3 & 28.8 & 9.7 & 40.6 \\
 & Uniform & 52.0 & 30.2 & 11.8 & 40.5 \\
 & Gaussian & 47.4 & 26.2 & 9.1 & 36.7 \\ \hline  \hline
\multirow{3}{*}{\textbf{Ours}}
 & Manual & {58.1} & {34.5} & {11.9} & {44.3} \\
 & Uniform & {55.6} & {32.3} & {12.3} & {42.9} \\
 & Gaussian & \textbf{58.2} & \textbf{35.9} & \textbf{12.8} & \textbf{44.8} \\
\bottomrule
\end{tabular}}
\label{tab:different label distributions}
\end{center}
\vspace{-10pt}
\end{table}


\begin{table}[t]
\small
\begin{center}
\caption{Statistical comparison of final localization results on THUMOS14. False Alarm, Precision, Recall, and F-measure are calculated under IoU threshold 0.5.
}
\setlength{\tabcolsep}{5.4pt}{
\begin{tabular}{c|c|c|c|c}
\toprule
{Method} & {False Alarm} & {Precision} & {Recall} & {F-measure} \\ \hline \hline
SF-Net~\cite{ma2020sf} & 73.3 & 26.7 & 51.8 & 35.1  \\
\textbf{Ours} & \textbf{59.5} & \textbf{40.5} & \textbf{57.6} & \textbf{47.4} \\
\bottomrule
\end{tabular}}
\label{tab:FP statistical indicators}
\end{center}
\vspace{-20pt}
\end{table}


\begin{table*}[t]
\small
\begin{center}
\caption{Comparison with the state-of-the-art methods on THUMOS14.
AVG(0.1-0.5) and AVG(0.3-0.7) are the average mAP from IoU 0.1 to 0.5 and from IoU 0.3 to 0.7, respectively.
The superscript ${\dagger}$ means that point-level labels are annotated manually. And the superscript ${\ddagger}$ indicates that point-level labels are simulated from ground truth boundary annotations.
}
\begin{tabular}{C{1.8cm}|C{1.9cm}|C{0.6cm}C{0.6cm}C{0.6cm}C{0.6cm}C{0.6cm}C{0.6cm}C{0.6cm}|C{1.2cm}|C{1.2cm}}
\toprule
\multirow{2}{*}{{Supervision}} & \multirow{2}{*}{{Method}} & \multicolumn{7}{c|}{{mAP@IoU}} & \multirow{2}{*}{{\begin{tabular}[c]{@{}c@{}}AVG\\ (0.1-0.5)\end{tabular}}} & \multirow{2}{*}{{\begin{tabular}[c]{@{}c@{}}AVG\\ (0.3-0.7)\end{tabular}}} \\ \cline{3-9}
 &  & {0.1} & {0.2} & {0.3} & {0.4} & {0.5} & {0.6} & {0.7} &  &  \\ \hline  \hline
\multirow{6}{*}{{Full}}
 & SSN~\cite{zhao2017temporal} & 66.0 & 59.4 & 51.9 & 41.0 & 29.8 & 19.6 & 10.7 & 49.62 & 30.60 \\
 & BSN~\cite{lin2018bsn} & - & - & 53.5 & 45.0 & 36.9 & 28.4 & 20.0 & - & 36.76 \\
 & AGCN~\cite{li2020graph} & 59.3 & 59.6 & 57.1 & 51.6 & 38.6 & 28.9 & 17.0 & 53.24 & 38.64 \\
 & A2Net~\cite{yang2020revisiting} & 61.1 & 60.2 & 58.6 & \textbf{54.1} & \textbf{45.5} & 32.5 & 17.2 & 55.90 & 41.58  \\
 & BUMR~\cite{zhao2020bottom} & 58.2 & 56.8 & 53.9 & 50.7 & 45.4 & \textbf{38.0} & \textbf{28.5} & 52.99 & \textbf{43.30}  \\  \cline{2-11}
 & \textbf{Ours} & \textbf{73.2}  & \textbf{65.6} & \textbf{60.1}  & 52.9 & 39.2 & 29.1 & 18.4 & \textbf{58.20}  & 39.94  \\
 \hline \hline
\multirow{8}{*}{{\begin{tabular}[c]{@{}c@{}}Weak\\Video-level\end{tabular}}}
 & STPN~\cite{nguyen2018weakly} & 45.3 & 38.8 & 31.1 & 23.5 & 16.2 & 9.8 & 5.1 & 30.98 & 17.14 \\
 & WTALC~\cite{paul2018w}  & 55.2 & 49.6 & 40.1 & 31.1 & 22.8 & 14.8 & 7.6 & 39.76 & 23.28 \\
 & CMCS~\cite{liu2019completeness} & 57.4 & 50.8 & 41.2 & 32.1 & 23.1 & 15.0 & 7.0 & 40.92 & 23.68 \\
 & BM~\cite{nguyen2019weakly} & 64.2 & 59.5 & 49.1 & 38.4 & 27.5 & 17.3 & 8.6 & 47.74 & 28.18  \\
 & BaSNet~\cite{lee2019background} & 58.2 & 52.3 & 44.6 & 36.0 & 27.0 & 18.6 & 10.4 & 43.64 & 29.81 \\
 & TSCN~\cite{zhai2020two} & 63.4 & 57.6 & 47.8 & 37.7 & 28.7 & 19.4 & 10.2 & 47.04  & 28.76 \\
 & DGAM~\cite{shi2020weakly} & 60.0 & 54.2 & 46.8 & 38.2 & 28.8 & 19.8 & 11.4 & 45.60 & 29.00 \\
 & A2CL~\cite{min2020adversarial} & 61.2 & 56.1 & 48.1 & 39.0 & 30.1 & 19.2 & 10.6 & 46.90 & 29.40 \\ \hline \hline
\multirow{2}{*}{{\begin{tabular}[c]{@{}c@{}}Weak\\Count-level \end{tabular}}}
 & STARN~\cite{xu2018segregated} & 68.8 & 60.0 & 48.7 & 34.7 & 23.0 & 11.7 & 6.2 & 47.04 & 24.86 \\
 & 3C-Net~\cite{narayan20193c} & 59.1 & 53.5 & 44.2 & 34.1 & 26.6 & - & 8.1 & 43.50 & -  \\
 \hline \hline
\multirow{5}{*}{{\begin{tabular}[c]{@{}c@{}}Weak\\Point-level\end{tabular}}}
 & SF-Net$^{\dagger}$~\cite{ma2020sf} & 71.0 & 63.4 & 53.2 & 40.7 & 29.3 & 18.4 & 9.6 & 51.52 & 30.24 \\
 & \textbf{Ours}$^{\dagger}$ & \textbf{72.8} & \textbf{64.9} & \textbf{58.1} & \textbf{46.4} & \textbf{34.5} & \textbf{21.8} & \textbf{11.9} & \textbf{55.34} & \textbf{34.54} \\  \cline{2-11}
 & ARST$^{\ddagger}$~\cite{moltisanti2019action} & 24.3 & 19.9 & 15.9 & 12.5 & 9.0 & - & - & 16.30 & - \\
 & SF-Net$^{\ddagger}$~\cite{ma2020sf} & 68.3 & 62.3 & 52.8 & 42.2 & 30.5 & 20.6 & 12.0 & 51.22 & 31.62 \\
 & \textbf{Ours}$^{\ddagger}$ & \textbf{72.3} & \textbf{64.7} & \textbf{58.2} & \textbf{47.1} & \textbf{35.9} & \textbf{23.0} & \textbf{12.8} & \textbf{55.64} & \textbf{35.40} \\
\bottomrule
\end{tabular}
\label{tab:THUMOS}
\end{center}
\end{table*}


\begin{table}[t]
\small
\begin{center}
\caption{Comparison on GTEA and BEOID. AVG denotes the average mAP at IoU thresholds 0.1:0.1:0.7.}
\setlength{\tabcolsep}{5.5pt}{
\begin{tabular}{c|c|cccc|c}
\toprule
\multirow{2}{*}{{Dataset}} & \multirow{2}{*}{{Method}} & \multicolumn{4}{c|}{{mAP@IoU}} & \multirow{2}{*}{{AVG}} \\ \cline{3-6}
 &  & {0.1} & {0.3} & {0.5} & {0.7} &  \\ \hline \hline
\multirow{5}{*}{\textbf{GTEA}}
 & SF~\cite{ma2020sf} & 50.0 & 35.6 & 21.6 & 17.7 & 30.5 \\
 & SFB~\cite{ma2020sf} & 52.9 & 34.9 & 17.2 & 11.0 & 28.0 \\
 & SFBA~\cite{ma2020sf} & 52.6 & 32.7 & 15.3 & 8.5 & 26.4 \\
 & SF-Net~\cite{ma2020sf} & 58.0 & 37.9 & 19.3 & 11.9 & 31.0 \\
 & \textbf{Ours} & \textbf{59.7} & \textbf{38.3} & \textbf{21.9} & \textbf{18.1} & \textbf{33.7} \\ \hline \hline
\multirow{5}{*}{\textbf{BEOID}}
 & SF~\cite{ma2020sf} & 54.1 & 24.1 & 6.7 & 1.5 & 19.7 \\
 & SFB~\cite{ma2020sf} & 57.2 & 26.8 & 9.3 & 1.7 & 21.7 \\
 & SFBA~\cite{ma2020sf} & 62.9 & 36.1 & 12.2 & 2.2 & 27.1 \\
 & SF-Net~\cite{ma2020sf} & 62.9 & 40.6 & 16.7 & 3.5 & 30.1 \\
 & \textbf{Ours} & \textbf{63.2} & \textbf{46.8} & \textbf{20.9} & \textbf{5.8} & \textbf{34.9} \\
\bottomrule
\end{tabular}}
\label{tab:GTEA and BEOID}
\end{center}
\vspace{-5pt}
\end{table}


\subsection{Ablation Studies and Comparisons}
\textbf{Ablation studies of the location prediction stage.}   \label{sec:Ablation Studies}
The foreground classification loss $L_{\mathrm{fg}}$, the background aware loss $L_{\mathrm{bg}}$ and the center offset ${\Delta}p$ are three important components. To investigate their contributions, we perform experiments on THUMOS14 and report the results in Table~\ref{tab:ablation studies}. (Without ${\Delta}p$, we treat the keypoint as the action center.)

Consistent with existing background modeling methods~\cite{lee2019background,nguyen2019weakly,lee2020background}, the background aware loss brings considerable improvement, with a gain of 6.4\% in mAP@0.5.
By modeling the auxiliary background category, our method is explicitly guided to distinguish actions from the background, resulting in more precise predictions.
Interestingly, predicting the center offset only leads to a slight improvement of 0.7\% in mAP@0.5.
We conjecture that it is because classification supervision focuses more on the discriminative action regions, while has limited guidance on the action center. Nevertheless, all components are effective and essential to achieve the best performance.

\begin{figure*}[t]
\begin{center}
\includegraphics [width=0.914\textwidth] {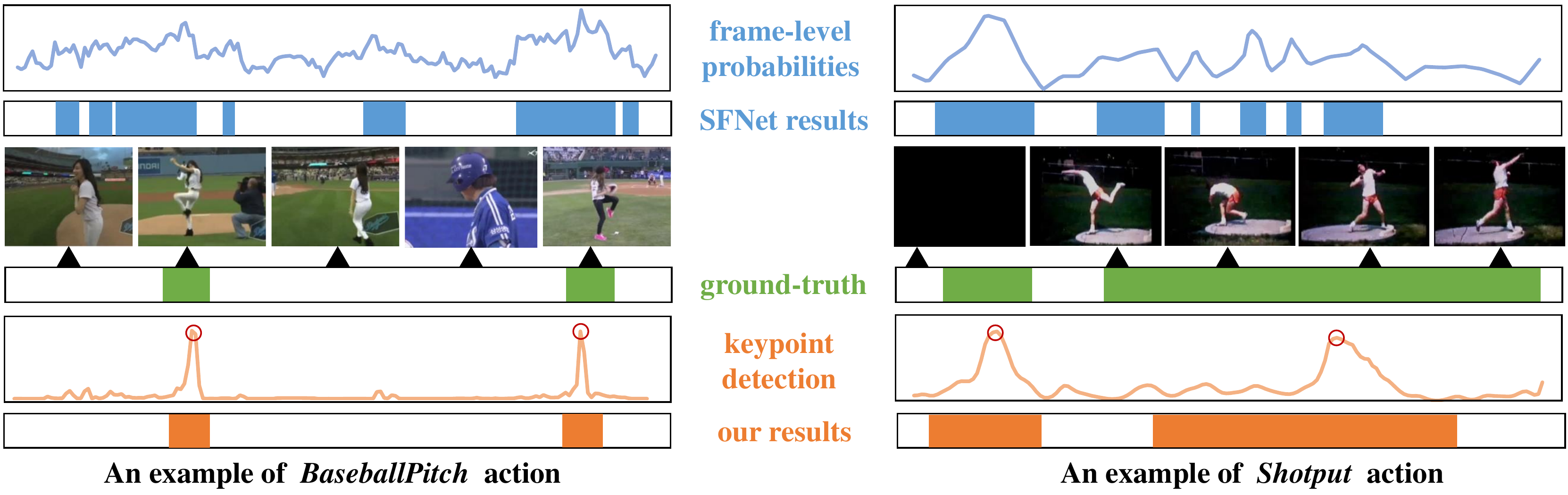}
\end{center}
\caption{
Qualitative comparison with SF-Net~\cite{ma2020sf} on THUMOS14. The first two rows are frame-level probabilities and localization results of SF-Net, the third row is the ground truth action intervals, and the last two rows are keypoint probabilities and localization results of our method.
\textbf{Left:} Our method detects keypoints as anchor points, effectively eliminating false positives in SF-Net.
\textbf{Right:} SF-Net suffers from discontinuous probabilities and gets scattered action fragments. While our method directly predicts action lengths and obtains more complete actions.
}
\label{fig:final results}
\end{figure*}

\textbf{Comparison of different label distributions.}
Due to some errors in manual point-level annotations, SF-Net~\cite{ma2020sf} explores to simulate labels via sampling on THUMOS14. For a comprehensive comparison, we follow the same settings to sample from existing ground-truth boundary labels through uniform distribution and Gaussian distribution, then obtain the simulated point-level annotations.

The detailed experimental results are shown in Table~\ref{tab:different label distributions}.
No matter which distribution is used, our method achieves gratifying results, revealing the generality of our method. Moreover, it can be found that our method performs best on Gaussian distribution and worst on uniform distribution, which is the opposite of SF-Net. We infer the reasons as follows. The proposal-based prediction paradigm makes our performance highly correlated with the action center. As discussed before, classification supervision only provides limited guidance to adjust the action center. Hence, keypoint detection plays a decisive role in finding the action center. Among the three distributions, Gaussian distribution can usually produce the sampling labels closest to the middle timestamps of actions, which causes the predicted keypoints nearest to action centers, and causally achieves the best performance in our method. By contrast, uniform distribution provides more temporal boundary information, which is more beneficial for the frame-level probability prediction paradigm, and more suitable for SF-Net.

\textbf{Comparison of statistics.}
To further confirm our efficacy, we collect some statistics based on the final localization results. As the existing state-of-the-art method in PTAL, SF-Net~\cite{ma2020sf} is chosen for comparison.
The results in Table~\ref{tab:FP statistical indicators} suggest that our method greatly reduces the false alarm and improves the precision accordingly, indicating that false positives are significantly suppressed. Moreover, a boost in recall demonstrates that our method effectively reduces omissions and detects more complete actions.

\subsection{Comparison with state-of-the-art methods}
Table~\ref{tab:THUMOS} compares our method with current state-of-the-art approaches on THUMOS14.
Using the same point-level annotations, our method outperforms other methods by a large margin regardless of the label distribution. It can be observed that there still exists a large performance gap of more than 10\% average mAP between previous point-supervised methods and fully-supervised methods. Benefits from the great reduction of the solution space, our method gains a substantial improvement and bridges the gap to 6\% average mAP.
Moreover, at some low IoU thresholds, our method is even comparable to several fully-supervised counterparts~\cite{shou2016temporal,gao2017turn,zhao2017temporal}. However, due to the lack of precise location supervision, our performance drops significantly as the IoU threshold increases.
Furthermore, our results under the fully-supervised setting are also demonstrated. In general, our method outperforms existing approaches at IoU thresholds 0.1, 0.2 and 0.3, while performs comparably to these methods at higher IoU thresholds.

We also present the quantitative comparison on GTEA and BEOID in Table~\ref{tab:GTEA and BEOID}.
SF, SFB and SFBA, which are three benchmark models designed in SF-Net~\cite{ma2020sf}, are included for comparison. On GTEA, our method achieves a new state-of-the-art performance, reaching 33.7\% average mAP. On BEOID, our method surpasses the best competitor by more than 4.5\% on the average mAP.

\subsection{Qualitative Results}
To demonstrate the superiority of our method intuitively, we visualize several qualitative results in Figure~\ref{fig:final results}. We also reproduce the results of SF-Net~\cite{ma2020sf} for better comparison.
As is evident, our method detects more precise and complete actions than SF-Net. More specifically, the frame-level probability curve of SF-Net has poor continuity and serious background noise, leading to inferior results. On the contrary, by detecting keypoints as action anchor points, our method effectively eliminates false positives. By directly predicting the center offset and action length, our method constrains continuity and obtains more complete results.

\section{Conclusions}
In this paper, we introduced the proposal-based prediction paradigm for PTAL. We trained a keypoint detector to discover action anchor points and rule out false positives. Based on each anchor point, we directly predicted the action length and center offset to form a proposal, ensuring the continuity of predictions.
To bridge this paradigm with weak supervision, we further designed a mapper module to transform the proposal location into a binary mask so that the model can be optimized by category labels.
Extensive Experiments on three benchmarks have verified the effectiveness and superior performance of our method.


{\small
\bibliographystyle{ieee_fullname}
\bibliography{PSTAL}
}

\end{document}